\documentclass[a4paper,fleqn]{cas-dc}

\usepackage[numbers]{natbib}
\usepackage{enumitem}
\usepackage{graphicx}
\usepackage{color}
\usepackage{bm}
\usepackage{amsfonts}
\usepackage{pifont}
\usepackage{stfloats}
\usepackage{amsmath, bm}
\usepackage{amssymb}
\usepackage{ragged2e} 
\usepackage{booktabs,makecell, multirow, tabularx}
\newcommand{\etal}{{\emph{et al. }}}
\usepackage{hyperref}
\usepackage{bbding}
\usepackage{algorithm}
\usepackage{algorithmic}

\usepackage{subfigure}
\newcommand{\ie}{{\emph{i.e., }}}
\newcommand{\eg}{{\emph{e.g., }}}
\newcommand{\vecc}[1]{{\boldsymbol{\mathbf{#1}}}}

\usepackage{tipa}

\newcommand{\Real}{\mathbb{R}}
\newcommand{\str}[1]{\vecc{#1}}  

\newcommand{\Lc}{\mathcal{L}}

\newcommand{\poincare}{Poincar\'e }
\newcommand{\lorentz}{\mathbb{L}^n_K}
\newcommand{\linner}[2]{\langle \mathbf{#1},\mathbf{#2}\rangle_\Lc}
\newcommand{\lnorm}[1]{\lVert #1\rVert_\Lc}
\newcommand{\tangent}[1]{\mathcal{T}_{\mathbf{#1}}\mathbb{L}^n_K}

\newcommand{\logmap}[2]{\log_#1^K(#2)}
\newcommand{\expmap}[2]{\exp_#1^K(#2)}
\newcommand{\riemanntensor}[1][x]{\mathfrak{g}_\mathbf{x}^K}

\def\tsc#1{\csdef{#1}{\textsc{\lowercase{#1}}\xspace}}
\tsc{WGM}
\tsc{QE}
\tsc{EP}
\tsc{PMS}
\tsc{BEC}
\tsc{DE}

\begin{document}
\let\WriteBookmarks\relax
\def\floatpagepagefraction{1}
\def\textpagefraction{.001}
\shorttitle{Learning Weakly Supervised Audio-Visual Violence Detection in Hyperbolic Space}
\shortauthors{Xiaogang Peng \etal}

\title [mode = title]{Learning Weakly Supervised Audio-Visual Violence Detection in Hyperbolic Space}                      

\author[1]{Xiaogang Peng}
\cormark[1]
\credit{Conceptualization, Methodology, Writing - Original Draft}
\author[2]{Hao Wen}
\cormark[1]
\credit{Investigation, Validation, Writing - Original Draft}
\author[1]{Yikai Luo}
\cormark[1]
\credit{Software, Validation, Data curation, Visualization}
\author[1]{Xiao Zhou}
\credit{Data curation, Visualization}
\author[1]{Keyang Yu}
\credit{Data curation, Visualization}
\author[1]{Ping Yang}
\credit{Writing - Review and Editing}
\author[1]{Zizhao Wu}
\cormark[2]
\ead{wuzizhao@hdu.edu.cn}
\credit{Resources, Writing - Review and Editing}

\address[1]{School of Digital Media and Art, Hangzhou Dianzi University, Hangzhou, China}
\address[2]{Academy for Engineering and Technology, National University of Defense Technology, China}

\cortext[0]{* indicates equal contribution}
\cortext[0]{** indicates corresponding author}

\begin{keywords}
    Weakly supervised learning\sep Hyperbolic space \sep Video violence detection
\end{keywords}
\maketitle

\begin{abstract}
 In recent years, the task of weakly supervised audio-visual violence detection has gained considerable attention. The goal of this task is to identify violent segments within multimodal data based on video-level labels. Despite advances in this field, traditional Euclidean neural networks, which have been used in prior research, encounter difficulties in capturing highly discriminative representations due to limitations of the feature space. To overcome this, we propose \textbf{HyperVD}, a novel framework that learns snippet embeddings in hyperbolic space to improve model discrimination. We contribute two branches of fully hyperbolic graph convolutional networks that excavate feature similarities and temporal relationships among snippets in hyperbolic space. By learning snippet representations in this space, the framework effectively learns semantic discrepancies between violent snippets and normal ones. Extensive experiments on the XD-Violence benchmark demonstrate that our method achieves 85.67\% AP, outperforming the state-of-the-art methods by a sizable margin.
\end{abstract}

\section{Introduction}

With the increase in the volume of digital content and the proliferation of social media platforms, automated violence detection has become increasingly important in various applications such as security and surveillance systems, crime prevention, and content moderation. However, annotating each frame in a video is a time-consuming and expensive process. To address this, current methods often utilize weakly supervised settings to formulate the problem as a multiple-instance learning (MIL) task \cite{c:1,c:2,c:3,c:4,c:5,c:6,c:8,c:13}. These methods treat a video as a bag of instances (\ie snippets or segments), and predict their labels based on the video-level annotations \cite{c:7}. 

\begin{figure}[h]
	\centering
	\includegraphics[width=0.45\textwidth]{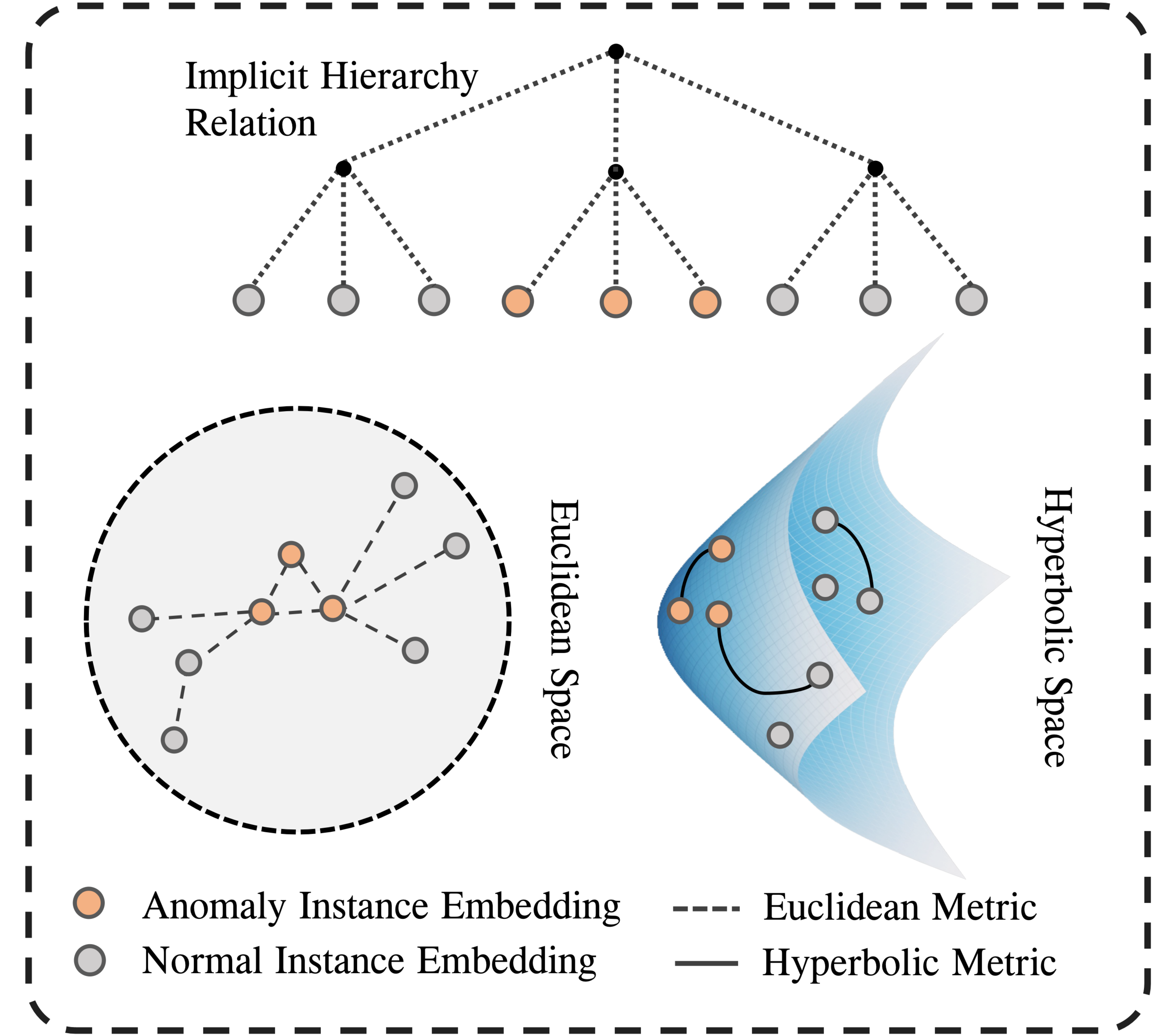} 
	\caption{Intuitively, there are implicit hierarchical relationships and substantial semantic discrepancies between violent instances and normal instances. These discrepancies can be difficult to capture using traditional Euclidean space methods, which may not be well-suited to represent complex hierarchical structures.}
	\label{fig1}
\end{figure}

Following the MIL paradigm, a number of weakly supervised violence detection methods have been proposed. For example, Zhu \etal\cite{Zhu_Newsam_2019} proposed a temporal augmented network to learn motion-aware features using attention blocks, while Tian \etal\cite{c:1} developed the Robust Temporal Feature Magnitude (RTFM) method to enhance model robustness through temporal attention and magnitude learning. Li \etal\cite{c:9} introduced a transformer-based framework and utilized multiple sequence learning to reduce the probability of selection errors. Furthermore, several multimodal approaches have been proposed, which jointly learn audio and visual representations to improve performance by leveraging complementary information from different modalities\cite{c:2,c:4,pang2021violence,Pu_Wu_2022}. For instance, Wu \etal\cite{c:2} proposed a GCN-based method to learn multimodal representations via graph learning, while Yu \etal \cite{c:4} presented a method that addresses modality asynchrony via modality-aware multiple instance learning.

Though the above-mentioned approaches have gained promising results, these multimodal methods may suffer heavy modality unbalance due to the presence of noise in audio signals collected from real-world scenarios. In this case, auditory modality contribute less than visual one for violence detection. In addition, previous methods have demonstrated the effectiveness of using graph representation learning to detect violent events by regarding each instance as a node in a graph \cite{c:2, c:3}, but they still struggle to differentiate violent and non-violent instances.

In this paper, we propose a new approach to address these limitations via graph representation learning. To our best knowledge, all the previous methods learn feature representation with deep neural networks in Euclidean space. However, graph-like data is proved to exhibit a highly non-Euclidean latent structure \cite{Bronstein2016GeometricDL, Ying2018HierarchicalGR} that challenges current Euclidean-based deep neural networks. As shown in Figure \ref{fig1}, there exist implicit hierarchical relationships and substantial semantic discrepancies between normal and violent instances, which are difficult to distinguish in Euclidean space. We argue that learning instance representations directly in a data-related space, such as hyperbolic manifolds, can favor the model discrimination, as it enables the model to capture and differentiate between subtle semantic differences that may hard to be explored in Euclidean space. 

Motivated by these findings, we propose a novel \textbf{HyperVD} framework based on the Lorentz model \cite{Nickel2018LearningCH} of hyperbolic geometry for weakly supervised audio-visual violence detection. Building the framework on hyperbolic geometry can benefit from the hyperbolic distance, which exponentially increases the distance between irrelevant samples compared to the distance between similar samples. In particular, our approach includes a detour fusion module to address the modality unbalance during the fusion stage, followed by projecting the fused embeddings of audio-visual features onto the hyperbolic manifold. Then we leverage two branches of fully hyperbolic graph convolutional networks to extract feature similarities and temporal relationships among instances in hyperbolic space. Furthermore, we concatenate the learned embeddings from the two branches and feed them into a hyperbolic classifier for violence prediction. To evaluate the effectiveness of our proposed approach, we conduct experiments on the XD-Violence dataset. Under weak supervision, our method can achieve the best performance of 85.67\% AP, outperforming the previous state-of-the-art method by 2.27\%. Extensive ablations also demonstrate the effectiveness of instance representation learning in hyperbolic space. 

In summary, the main contributions are stated as follows:
\begin{itemize}
	\item[$\bullet$]We analyze the weakness of learning instance representations using traditional Euclidean-based methods and present a novel HyperVD framework to effectively explore the instances' semantic discrepancy for weakly supervised violence detection via hyperbolic geometry, leading to more powerful discrimination.
	\item[$\bullet$]Experimental results show our framework outperforms the state-of-the-art methods on the XD-Violence dataset. The ablation study further gives insights into how each proposed component contributes to the success of the model.
\end{itemize}

\begin{figure*}[t]
	\centering
	\includegraphics[width=1.0\textwidth]{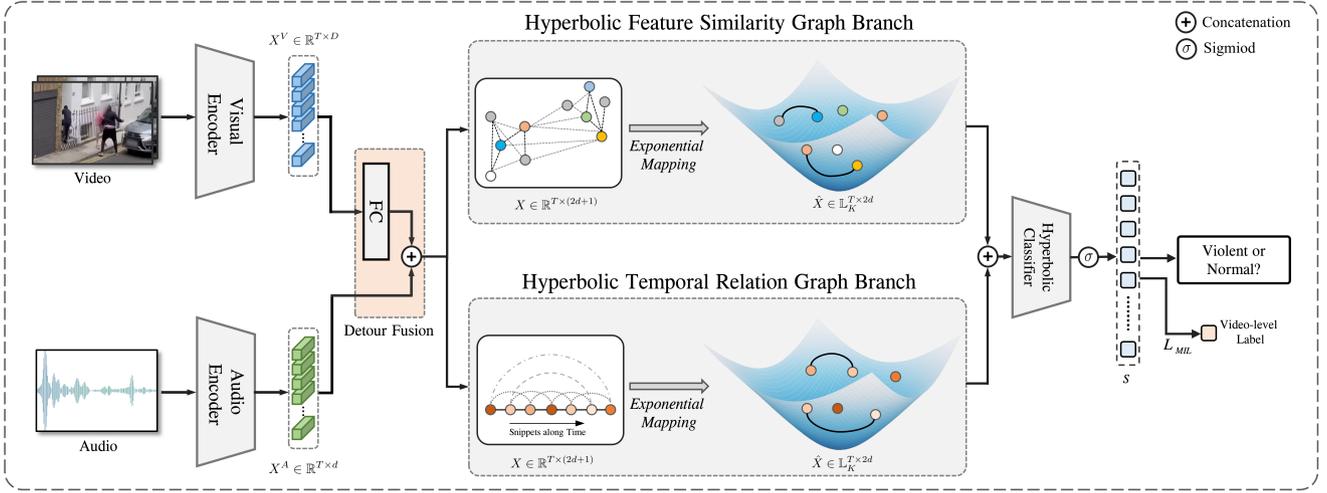} 
	\caption{Overview of our HyperVD framework. Our approach consists of fours parts: detour fusion, hyperbolic feature similarity graph branch, hyperbolic temporal relation graph branch and hyperbolic classifier. Taking audio and visual features extracted from pretrained networks as inputs, we design a simple yet effective module to fuse audio-visual information. Then two hyperbolic graph branches learn instance representations via feature similarity and temporal relation in hyperbolic space. Finally, a hyperbolic classifier is deployed to predict violent scores for each instance. The entire framework is trained jointly in a weakly supervised manner, and we adopt the multiple instance learning (MIL) strategy for optimization.}
	\label{fig2}
\end{figure*}

\section{Related Works}
\noindent\textbf{Weakly Supervised Violence Detection.}
Weakly supervised violence detection aims to identify violent segments in videos by utilizing video-level labels. Since the publication of the first paper~\cite{Ding2014ViolenceDI} utilizing deep learning methods, the field of violence detection has made tremendous strides. To eliminate irrelevant information and enhance the accuracy of detection, the MIL~\cite{maron1997framework} framework is widely employed in this process. Most existing works~\cite{RendnSegador2023CrimeNetNS, bermejo2011violence, deniz2014fast, feng2021mist, peixoto2019toward, ristea2021self, c:8, zhang2019temporal, zhang2016new, xu2014video} consider violence detection as solely a visual task, and CNN-based networks are utilized to encode visual features.
Sultani \etal \cite{c:7} propose a MIL ranking loss with sparsity and smoothness constraints for deep learning networks to learn the anomaly scores in video segments. Li \etal \cite{c:9} develope a multi-sequence learning model based on Transformer \cite{Vaswani2017AttentionIA} to reduce the probability of selection errors.  
A recent research \cite{c:2} releases a large-scale audio-visual violence dataset. To facilitate inter-modality interactions, Yu \etal \cite{c:4} propose a lightweight two-stream network and utilize modality-aware contrast and self-distillation to achieve discriminative multimodal learning. To focus on the implication of normal data, Zhou \etal \cite{c:10} propose an dual memory units module with uncertainty regulation to learn both the representations of normal data and the discriminative features of abnormal data. Different from prior methods, we project the fused embeddings of audio-visual features on the hyperbolic manifold, and employ fully hyperbolic graph convolutional networks to effectively excavate the semantic discrepancy between violent and non-violent instances.

\noindent\textbf{Neural Networks in Hyperbolic Space.}
Hyperbolic space is a kind of non-Euclidean space with constant negative Gaussian curvature. Recently, hyperbolic space has been drawing increasing interest in machine learning and neural information science due to its appealing properties in representing data with hidden hierarchies \cite{Nickel2017PoincarEF, Sala2018RepresentationTF, Nickel2018LearningCH, Wang2019HyperbolicHI}. Nickel et al. \cite{Nickel2017PoincarEF} conduct a groundbreaking study of learning representation in hyperbolic spaces using the Poincar{\'e} ball model. Sala et al. \cite{Sala2018RepresentationTF} analyze the trade-offs of embedding size and numerical precision in these different models and Ganea et al. \cite{Ganea2018HyperbolicEC} extend these methods to undirected graphs. On this basis, Ganea et al. \cite{Ganea2018HyperbolicNN} define a hyperbolic neural network, which bridges the gap between hyperbolic space and deep learning. Nickel et al. \cite{Nickel2018LearningCH} and Wilson et al. \cite{Wilson2018GradientDI} demonstrate that using the Lorentzian model of hyperbolic space can result in more efficient and simpler optimizers compared to the Poincaré ball. In recent research \cite{gu2019learning}, neural networks have been developed based on Cartesian products of isotropic spaces. In fact, hyperbolic space has been well incorporated into recent advanced deep learning models such as the recurrent neural network \cite{Ganea2018HyperbolicNN}, graph neural network \cite{Liu2019HyperbolicGN}, and attention network \cite{Glehre2019HyperbolicAN}. Based on these studies for deep learning paradigms, we investigate the effectiveness of learning weakly-supervised audio-visual violence detection in hyperbolic space using hyperbolic neural networks.

\section{Preliminaries}
Before describing our method's details, in this section, we will introduce background knowledge of the hyperbolic geometry with its modeling, \ie Lorentz model, and the hyperbolic graph convolutional networks that we adopt in this work.

\noindent\textbf{Hyperbolic Geometry.}
\label{subsec:3_1}
Hyperbolic geomerty is a non-Euclidean geometry with a constant negative curvature $K$. The hyperbolic geometry models have been applied in previous studies: the \poincare ball (\poincare disk) \cite{ganea2018hyperbolic}, the \poincare half-plane model \cite{tifrea2018poincar}, the Klein model \cite{gulcehre2018hyperbolic}, and the Lorentz (Hyperboloid) model \cite{Nickel2018LearningCH}. We select the Lorentz model as the framework base, considering the numerical stability and calculation simplicity of its exponential and logarithmic maps and distance functions.

We denote $\lorentz=(\Lc^n, \riemanntensor)$ as an n-dimensional Lorentz model with constant negative curvature K. $\Lc^n$ is a point set satisfying:
\small{
\begin{align}
	\label{eq1}
	\Lc^n:=\{{x\in\mathbb{R}^{n+1}: <x,x>_{\Lc}=\frac{1}{K}, x_i>0}\}.
\end{align}
}
The Lorentzian scalar product is defined as:
\small{
\begin{align}
	\label{eq2}
	<x, y>_{\Lc}:=-x_0y_0+\sum_{i=1}^{n}{x_iy_i}, 
\end{align}
}
where $\Lc^n$ is the upper sheet of hyperboloid in an (n+1)-dimensional Minkowski space with the origin $(\sqrt{-1/K}, 0,...,0)$. For simplicity, we denote point $x$ in the Lorentz model as $x\in\mathbb{L}^n_K$.\\

\noindent\textit{Tangent Space.} The tangent space at $x$ is defined as an n-dimensional vector space approximating $\lorentz$ around $x$,
\begin{align}
	\label{eq3}
	\tangent{x}:=\{\mathbf{y}\in \Real^{n+1} \mid \linner{y}{x}=0\}.
\end{align}
Note that $\tangent{x}$ is a Euclidean subspace of $\mathbb{R}^{n+1}$.\\

\noindent\textit{Exponential and Logarithmic Maps.} The mapping of points between the hyperbolic space $\lorentz$ and the Euclidean subspace $\tangent{x}$ can be done by exponential map and logarithmic map. The exponential map can map any tangent vector $z\in\tangent{x}$ to $\lorentz$, and the logarithmic map is a reverse map that maps back to the tangent space. These two maps can be written as:
\begin{align}
	\label{eq4}
	\expmap{x}{z} &= \cosh(\sqrt{-K}\lVert\mathbf{z}\rVert_\Lc)\mathbf{x} + \sinh(\sqrt{-K}\lVert\mathbf{z}\rVert_\Lc)\frac{\mathbf{z}}{\sqrt{-K}\lVert\mathbf{z}\rVert_\Lc},\\
	\label{eq5}
	\logmap{\mathbf{x}}{\mathbf{y}} &= d_{\mathbb{L}}^{K}(\mathbf{x},\mathbf{y})\frac{\mathbf{y}-K<\mathbf{x},\mathbf{y}>_{\mathcal{L}}}{\lVert \mathbf{y}-K<\mathbf{x},\mathbf{y}>_{\mathcal{L}}\rVert _{\mathcal{L}}},
\end{align}
where $\lVert\mathbf{z}\rVert_\Lc = \sqrt{\linner{z}{z}}$ denotes Lorentzian norm of $\mathbf{z}$ and $ d_{\mathbb{L}}^{K}(\cdot,\cdot)$ denotes the Lorentzian intrinsic distance function between two points $\mathbf{x},\mathbf{y}\in \lorentz$, which is given as:
\begin{align}
	\label{eq6}
	d_{\mathbb{L}}^{K}(\mathbf{x},\mathbf{y}) = arccosh(K<\mathbf{x},\mathbf{y}>_{\mathcal{L}}).
\end{align}

\subsection{Hyperbolic Graph Convolutional Networks}
Recently, several hyperbolic GCNs have been proposed to extend Euclidean graph convolution to the hyperboloid model and have obtained promising results in a wide range of scenarios\cite{c:14}. In order to adapt widely-used Euclidean neural operations, such as matrix-vector multiplication, to hyperbolic spaces, existing methods formalize most of the operation in a hybrid way that involves transforming features between hyperbolic spaces and tangent spaces using logarithmic and exponential maps, and performing neural operations in tangent spaces. For instance, in HGCN \cite{c:13}, let $h_{i,K}^{n} \in\mathbb{H}_{K}^{n}$ be a $n$-dimensional node features of node $i$ on hyperboloid manifold $\mathbb{H}_{K}^{n}$, $N(i)$ be a set of its neighborhoods with adjacent matrix $A_{ij}$, and $W$ be a weight matrix. Its message passing rules consist of \emph{feature transformation}:
\begin{align}
	\label{eq7}
	h_{i,K}^{d} = {\rm exp}_{0}^{K}({   W\logmap{0}{h_{i,K}^n}   }),
\end{align}
and \emph{neighborhood aggregation}:
\begin{align}
	\label{eq8}
	\rm{Agg}(h_{i,K}^d) &= {\rm exp}_{h_i}^{K}({\sum_{j\in N(i)\cup{i}} {A_{ij} \logmap{{h_i}}{{h_{i,K}^d}} }} ),
\end{align}
where $\expmap{0}{\cdot}$ and $\logmap{0}{\cdot}$ are logarithmic and exponential maps of the $\mathbb{H}_{K}^{n}$. The above hybrid manner does not fully satisfy hyperbolic geometry, causing distortion for the node features of graphs and weakening the stability of models \cite{c:15, c:16}. 

Therefore, Chen \etal \cite{c:16} proposed a fully hyperbolic neural network based on Lorentz model by adapting the Lorentz transformations (including boost and rotation) to formalize essential neural operations and proved that linear transformation in the tangent space at the origin of hyperbolic spaces is equivalent to performing a Lorentz rotation with relaxed restrictions. Readers could refer to \cite{c:16} for more detailed derivation. For simplicity, they provide a more general formula \footnote{This general formula is no longer fully hyperbolic. It is a relaxation in implementation, while the input and output are still guaranteed to lie in the Lorentz model \cite{c:16}.} of their hyperbolic linear layer for \emph{feature transformation} with activation, dropout, bias and normalization,
\begin{align}
	\label{eq9}
	\mathbf{y} = {\rm HL}(\mathbf{x}) = \left[\begin{smallmatrix}
		\sqrt{\lVert \phi(\mathbf{Wx, v}) \rVert^2 - 1/K}  \\
		\phi(\mathbf{W}\mathbf{x}, \mathbf{v})
	\end{smallmatrix}\right], 
\end{align}
where $\mathbf{x} \in\lorentz$, $\mathbf{W}\in\mathbb{R}^{d\times(n+1)}$, $\mathbf{v} \in \mathbb{R}^{n+1}$ denotes a velocity (ratio to the speed of light) in the Lorentz transformations, and $\phi$ is an operation function: for the dropout, the function is $\phi(\mathbf{Wx, v}) = \mathbf{W}{Dropout}(\mathbf{x})$; for the activation and normalization $\phi(\mathbf{Wx, v}) = \frac{\lambda \sigma(\mathbf{v}^\intercal\mathbf{x}+b')}{\lVert \mathbf{W}h(\mathbf{x})+b \rVert}(\mathbf{W}h(\mathbf{x})+b)$, where $\sigma$ is the sigmoid function, $b$ and $b'$ are bias terms, $\lambda > 0$ controls the scaling range, $h$ is the activation function. Further, their proposed \emph{neighborhood aggregation} can be defined as:
\begin{align}
	\label{eq10}
	{\rm HyperAgg}(\mathbf{y}_i) = \frac{\sum_{j=1}^{m}A_{ij}\mathbf{y}_j}{\sqrt{-\textit{K}} \big | \lnorm{\sum_{k=1}^{m}A_{ik}\mathbf{y}_k} \big|} ,
\end{align}
where $\rm{m}$ is the number of points. The non-linear activation of this method is omitted in the last operation for it is already integrated into the hyperbolic linear layer. In our study, we adapt the fully hyperbolic graph convolutional network into our framework to explore the efficacy of instance representation learning in hyperbolic space.

\section{Method}
In this section, we first define the formulation and problem statement. Then we introduce our proposed framework in detail, which mainly consists of four parts: detour fusion, hyperbolic feature similarity graph branch, hyperbolic temporal relation graph branch and hyperbolic classifier. The illustration of the framework is shown in Figure \ref{fig2}.

\subsection{Formulation and Problem Statement}
Given an audio-visual video sequence $M=\{M^V_i, M^A_i\}_{i=1}^T$ with $T$ non-overlapping multimodal segments, where each segment contains 16 frames, and $M^V_i$ and $M^V_i$ denotes visual and audio segment, respectively. The annotated video-level label $Y \in \{1, 0\}$ indicates whether a violent event exists in this video. To avoid additional training overhead, we utilize the well-trained backbones (I3D\cite{c:12} and VGGish \cite{Gemmeke_Ellis, Hershey_Chaudhuri}) to extract visual features $X^V \in{\mathbb{R}^{T\times D}}$ and audio features $X^A \in{\mathbb{R}^{T\times d}}$, respectively, where $D$ and $d$ are the feature dimensions. Like prior works \cite{c:2, c:4, c:7, c:9}, our method aims to employ the multiple instance learning (MIL) procedure to distinguish whether it contains violent events (instances) in a weakly-supervised manner, utilizing just video-level labels $Y$ for optimization.

\subsection{Multimodal Fusion}
Here we discuss several commonly-used multimodal fusion manners in the early and middle stages for comparative experiments.

\noindent\textbf{Concat Fusion.}
A straightforward approach is to simply concatenate all the features of both modalities and then fuse them via fully-connected layers (FC). The output $X$ of the concat fusion scheme can be expressed as $X = f(X^A \oplus X^V)$, where $f(\cdot)$ is two-layered FC and $\oplus$ is concatenation operation.

\noindent\textbf{Additive Fusion.}
We combine the information from both modalities using component-wise addition, \ie $X = f_a(X^{A}) + f_v(X^{V})$, where $f_a(\cdot)$ and $f_v(\cdot)$ are two corresponding FC to keep the dimension of input features identical.  

\noindent\textbf{Gated Fusion.}
We investigate a gated fusion method proposed in \cite{kiela2018efficient}, which allows one modality to “gate” or “attend” over the other modality, via a sigmoid non-linearity, \ie $X = W(UX^A * VX^V)$, where $U$,$V$, and $W$ are weight matrices. One can think of this approach as performing attention from one modality over the other. 

\noindent\textbf{Bilinear \& Concat.}
We utilize two linear layers for both input features of two modalities and keep their dimension identical, followed by a concatenation operation, \ie $X = UX^A \oplus VX^V$, where $U$ and $V$ are weight matrices.

\noindent\textbf{Our Detour Fusion}
Let $X^V$ and $X^A$ denote the auditory and visual features extracted by the backbones, and $X = \{x_i\}_{i=1}^T$ denote the fusion of the features from the two modalities. 

In audio-visual violence detection, there is a distinctive modality imbalance between auditory and visual signals, unlike other typical multimodal tasks. Audio signals are frequently affected by noise stemming from the capture device source, which can degrade their quality. On the other hand, visual signals tend to be more informative and reliable, making them crucial for effective violence detection. Based on this intuition, the visual modality may be expected to contribute more to violence detection, compared to the auditory modality. Therefore, we utilize a simple and efficient detour fusion manner that only feeds visual features into FC layers, ensuring that the visual features have the same dimension as the audio ones. Then, we concatenate the visual and audio features to form a joint representation, denoted as $X = f_v(X^V) \oplus X^A$, where $f_v$ is a two-layered FC and $X \in \mathbb{R}^{T\times 2d}$. To a certain extent, this detour operation can give more importance to the visual modality than the audio modality. The experimental results validate the effectiveness of our detour fusion method, outperforming other commonly used fusion techniques. The implementation details of other fusion methods can be found in the Appendix.

\subsection{HFSG Branch}
Prior works have shown promising power of GCNs for video understanding \cite{Wang_Gupta_2018, c:3, Zeng_Huang_Tan_Rong_Zhao_Huang_Gan_2019, c:2}. Here, we leverage the fully hyperbolic GCN to learn discriminative representations via hyperbolic geometry. We first project fused features $X$ into hyperbolic space by exponential map $\expmap{x}{\cdot}$ and have $\hat{X}\in \mathbb{L}_{K}^{T\times 2d}$. Then we define adjacent matrix $A^{\mathbb{L}} \in \mathbb{R}^{T \times T}$ via hyperbolic feature similarity:
\begin{align}
	\label{eq11}
	A_{ij}^{\mathbb{L}} = softmax(g(\hat{x}_i, \hat{x}_j)),\\
	\label{eq12}
	g(\hat{x}_i, \hat{x}_j)= exp(-d_{\mathbb{L}}^{K}(\hat{x}_i, \hat{x}_j)),
\end{align}
where the element $A_{ij}^{\mathbb{L}}$ measures the hyperbolic feature similarity between the $i$th and $j$th snippets via Lorentzian intrinsic distance $d_{\mathbb{L}}^{K}(\cdot,\cdot)$ instead of cosine similarity or other Euclidean metrics. Since an adjacency matrix should be non-negative, we bound the similarity to the range (0, 1] with an exponential function $exp(\cdot)$. Before $softmax$ normalization, we also employ the thresholding operation to eliminate weak relations and strengthen correlations of more similar pairs in hyperbolic space. The thresholding can be defined as:
\begin{equation}
	\label{eq13}
	g(\hat{x}_i, \hat{x}_j) = \left\{
	\begin{aligned}
		g(\hat{x}_i, \hat{x}_j), \quad g(\hat{x}_i, \hat{x}_j)\textgreater\tau \\
		0, \quad g(\hat{x}_i, \hat{x}_j)\leq\tau \\ 
	\end{aligned}
	\right.
\end{equation}
where $\tau$ is the threshold value.

Given the hyperbolic embeddings $\hat{X}$, we leverage the hyperbolic linear layer $\rm{HL}(\cdot)$ for \emph{feature transformation}, which incorporates an activation layer for non-linear activation, followed by \emph{neighborhood aggregation} $\rm{HyperAgg}$ as elaborated in equation \ref{eq10}. The overall operations are as follows: 
\begin{align}
	\label{eq14}
	\hat{x}_{i}^{l} = \frac{\sum_{j=1}^{T}A_{ij}^{\mathbb{L}} {\rm HL}( \hat{x}_{i}^{l-1}) }{\sqrt{-\textit{K}} \big | \lnorm{\sum_{k=1}^{T}A_{ik}^{\mathbb{L}} {\rm HL}(\hat{x}_{i}^{l-1}) } \big|}  ,   
\end{align}
where $\hat{x}_{i}^{l}$ refers to the hyperbolic representation of the $i$th snippet at the layer $l$. The output of this branch is computed as:
\begin{align}
	\label{eq15}
	\hat{X}^{\mathbb{L}} = Dropout(LeakyReLU(\hat{X}^{l+1})).
\end{align}

\subsection{HTRG Branch}
Although the hyperbolic feature similarity branch can capture long-range dependencies by measuring the similarity of snippets between any two positions, irrespective of their temporal position information, the temporal relation is also crucial for numerous video-based tasks. To address this issue, we construct a temporal relation graph directly based on the temporal structure of a video and learn the temporal relation among snippets in hyperbolic space. Its adjacency matrix $A^{\mathbb{T}} \in \mathbb{R}^{T \times T} $is only dependent on temporal positions of the $i$th and $j$th snippets, which can be defined as:
\begin{align}
	\label{eq16}
	A^{\mathbb{T}}_{ij} = exp(- \lVert i-j\rVert^{\gamma}),
\end{align}
where $\gamma$ is a hyper-parameter that controls the scope of temporal distance.

Likewise, we obtain hyperbolic embeddings via $\hat{X}=\expmap{x}{\emph{X}}$, and forward $\hat{X}$ and $A^{\mathbb{T}}$ into the hyperbolic GCN to learn temporal relationships in hyperbolic space via equation \ref{eq14}. The final output is also computed as:
\begin{align}
	\label{eq17}
	\hat{X}^{\mathbb{T}} = Dropout(LeakyReLU(\hat{X}^{l+1})).
\end{align}

\subsection{Hyperbolic Classifier}
The output embeddings of the two branches still reside on the hyperbolic manifold, where it is not feasible to directly classify using a Euclidean-based linear layer. As shown in Figure \ref{fig2}, to predict violent scores $S\in \mathbb{R}^{T\times 1}$, we concatenate the embeddings and input them into a hyperbolic classifier, which can be formalized as:
\begin{align}
	\label{eq18}
	S = \sigma((\epsilon + \epsilon <\hat{X}^{\mathbb{L}} \oplus \hat{X}^{\mathbb{T}}, W>_{\mathcal{L}}) + b),
\end{align}
where $\sigma$ is sigmoid function and $W$ is weight matrices. $b$ and $\epsilon$ denotes bias term and hyper-parameter, respectively.

\subsection{Objective Function}
In this paper, violent detection is treated as a MIL task under weak supervision. Following \cite{c:2,c:7}, we use the mean value of the $k$-max predictive scores in a video bag as the violent score, where $k = \lfloor \frac{T}{q} + 1\rfloor $. High scoring $k$-max predictions in the positive bag are more likely to include violent events, whereas the $k$-max predictions in the negative bag are typically hard samples. Consequently, the objective function is as follows:
\begin{equation}
	\small{
		L_{MIL} = \frac{1}{N}\sum_{i=1}^{N}{-Y_{i}log(\Bar{S})},
	}
\end{equation}
where $\Bar{S}$ is the average value of the $k$-max predictions in the video bag and $Y_i$ is the binary video-level annotation.

\begin{table}[t]
	\caption{Comparison of the frame-level AP performance on XD-Violence. Bold numbers indicate the best performances. The methods with † and * are re-implemented and reported by \cite{c:4}. The top performance is highlighted in bold, while the second-best performance is highlighted by underlining.}
	\resizebox{1.0\columnwidth}{!}{
		\begin{tabular}{c|c|c|c|c}
			\hline
			Manner                  &  Method                       & Modality      & AP(\%)         & Param.(M)\\ \hline
			\multirow{3}{*}{Unsup}   & SVM baseline                 &   -          & 50.78           &  -     \\
			&  OCSVM              &   -          & 27.25            & -  \\
			& Hasan \etal         &   -          & 30.77           & -  \\ \hline
			\multirow{6}{*}{W.Sup}& Sultani \etal† (2018)  & V            & 75.68           & -  \\ 
			& Wu \etal (2021)     & V            & 75.90            & -  \\
			& RTFM (2021)         & V            & 77.81            & 12.067 \\
   			& MSL \etal (2022)     & V           & 78.28            & - \\            
			& S3R (2022)         & V            & 80.26            & -  \\       
   			& UR-DMU (2023)      &  V       &  81.66           & -     \\
                & Zhang \etal (2023) & V       &  78.74            & -     \\
   			& Wu \etal (2020)         & A + V            & 78.64     & 0.843  \\
			& Wu \etal† (2020)         & A + V            & 78.66       & 1.539  \\ 
			& RTFM* (2021)        & A + V        & 78.54           & 13.510 \\
			& RTFM† (2021)        & A + V        & 78.54           & 13.190 \\
			& Pang \etal (2021)   & A + V       & 81.69            & 1.876 \\  
			& MACIL-SD (2022)        & A + V       & \underline{83.40}            & \underline{0.678} \\ 
			& UR-DMU (2023)      &  A+ V       &  81.77           & -     \\
			& Zhang \etal (2023) & A + V       &  81.43            & -     \\\hline
			\multirow{2}{*}{W.Sup}& HyperVD (ours)                           &   V        & \textbf{82.51}   & \textbf{0.599}\\ 
                & HyperVD (ours)                           &  A + V        & \textbf{85.67}   & \textbf{0.607} \\ \hline
			
	\end{tabular}}
	\label{table1}
\end{table}

\section{Experiments}

\subsection{Implementation Details}
\noindent\textbf{Feature Extraction.} To make a fair comparison, we employ the same procedure for feature extraction as previous methods \cite{c:1,c:2, pang2021violence,c:4}. Specifically, to extract visual features, we use the I3D network \cite{Carreira_Zisserman_2017} pretrained on the Kinetics-400 dataset. For audio features, we employ the VGGish network \cite{Gemmeke_Ellis, Hershey_Chaudhuri}, which was pretrained on a large dataset of YouTube videos. Visual features are extracted at a sample rate of 24 frames per second, using a sliding window approach with a window size of 16 frames. For the auditory data, we divide each audio recording into 960-millisecond segments with overlap, and then compute the log-mel spectrogram using a resolution of 96 x 64 bins. This allows us to extract rich and informative auditory features that can be combined with the visual features to enhance the performance of our violence detection model.


\noindent\textbf{HyperVD Architecture and Settings.} For the detour fusion module, we apply two 1D convolutional layers with LeakyReLU activation and dropout to learn the visual features. In the hyperbolic space, we utilize two hyperbolic graph convolutional layers for the HSFG and HTRG branches. The input dimensions for both branches are 257, and the hidden dimensions are set to 32. The negative curvature constant, denoted as $K$, is a fixed value of -1.

\noindent\textbf{Training Details.} The entire network is trained on an NVIDIA RTX 3090 GPU for 50 epochs. We set the batch size as 128 during training, and set the initial learning rate as 5e-4, which is dynamically adjusted by a cosine annealing scheduler. For hyper-parameters, we set $\gamma$ as 1, $\epsilon$ as 2, and dropout rate as 0.6. We use Adam as the optimizer without weight decay. For the MIL, we set the value $k$ of $k$-max activation as $\lfloor \frac{T}{16} + 1\rfloor$, where $T$ denotes the length of input feature.

\begin{table}[t]
\centering
	\caption{Ablation studies for different multimodal fusion manners. The method with * is re-implemented by us by replacing its original concat fusion with our detour fusion.}
	\label{table2}
	\resizebox{0.95\columnwidth}{!}{\begin{tabular}{cclc}
		\toprule
		Index    & Manner  & AP(\%) & Param.(M)\\
		\midrule 
		1 &     Wu \etal* (2020)  & 79.86 \textcolor{red}{($\uparrow1.22$)} & 0.851  \\\midrule 
		2 &      Concat Fusion &	83.35	& 0.758\\
		3 &     Additive Fusion &	82.41	& 0.594 \\
		4 &      Gated Fusion	&   82.51	& 0.657  \\
		5 &     Bilinear \& Concat	& 81.33	& 0.644\\
		6 &     Detour Fusion (ours)& 85.67 	& 0.607 \\ 
		\bottomrule
	\end{tabular}}
\end{table}

\begin{figure*}[t]
	\centering
	\includegraphics[width=1.0\textwidth]{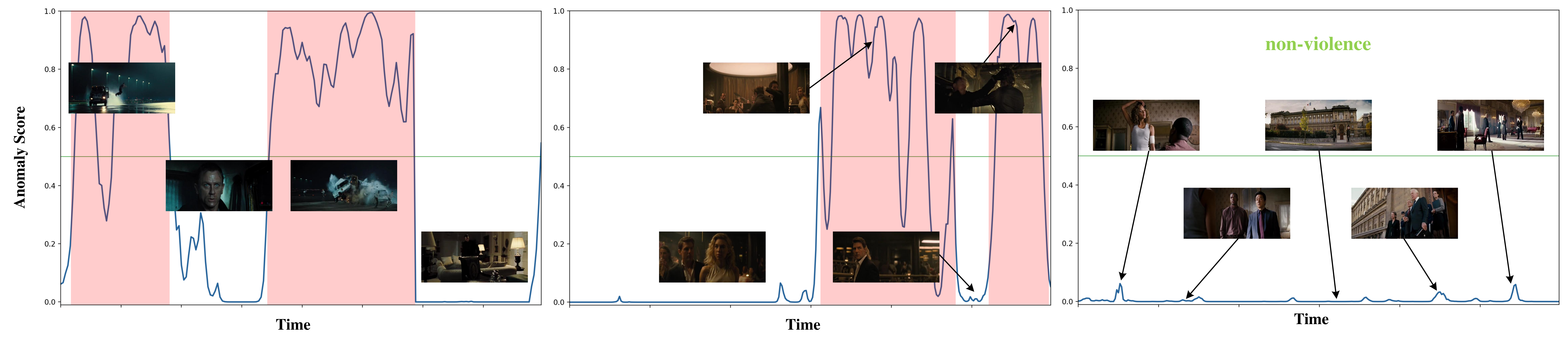} 
	\caption{Visualization of anomaly score curves. The horizontal axis represents the time, and the vertical axis represents the anomaly scores. The first row includes two samples of videos containing violent events, and the second row includes samples from normal videos. The blue curves indicate the predicted abnormal scores of the video frames, and the red areas indicate the locations of abnormal events.}
	\label{fig3}
\end{figure*}

\subsection{Dataset}
XD-Violence \cite{c:2} is a recently released large-scale audio-visual violence detection dataset, compiled from real-world movies, web videos, sport streaming, security cameras, and CCTVs. This dataset contains 4754 untrimmed films with video-level labels in the training set and frame-level labels in the testing set, for a total runtime of nearly 217 hours. Following \cite{c:2, pang2021violence, c:4}, we select this XD-Violence dataset as our benchmark to verify the efficiency of our proposed multimodal framework.
During inference, we use the Average Precision (AP) metric for evaluation following previous works \cite{c:1,c:2, pang2021violence,c:4}. It is important to note that higher values of AP correspond to better performance on the dataset.

\begin{table}[t]
\centering
	\caption{Ablation studies for utilizing various GCNs with different geometry models and different feature similarity metrics. $\mathbb{E}$, $\mathbb{B}$ and $\mathbb{L}$ indicate Euclidean, \poincare and Lorentz model, respectively.}
	\label{table3}
	\resizebox{0.9\columnwidth}{!}{\begin{tabular}{ccccc}
		\toprule
		Index  & Network & Model & Feature Similarity & AP(\%)\\
		\midrule
		1 &  GCN & $\mathbb{E}$ & Cosine Similarity & 79.85\\
		2 & HGCN & $\mathbb{B}$ & Cosine Similarity & 81.62\\
		3 & HGCN & $\mathbb{B}$ & \poincare Distance & 82.88 \\ \hline
		4  &  FHGCN & $\mathbb{L}$ & Cosine Similarity & 83.25\\
		5 & FHGCN & $\mathbb{L}$ & Lorentzian Distance & \textbf{85.67}\\
		\bottomrule
	\end{tabular}}
\end{table}
\vspace{-1mm}

\subsection{Quantitative Results}
We compare our proposed approach with previous state-of-the-art methods, including (1) unsupervised methods: SVM baseline, OCSVM\cite{c:5}, and Hasan \etal\cite{c:6}; (2) unimodal weakly-supervised methods: Sultani \etal \cite{c:7}, Wu \etal \cite{c:8} RTFM \cite{c:1},  MSL \cite{c:9}, S3R \cite{c:11}, UR-DMU \cite{c:10} and Zhang \etal\cite{zhang2022exploiting}; (3) audio-visual weakly-supervised method: Wu \etal \cite{c:2},  Pang \etal \cite{pang2021violence}, MACIL-SD \cite{c:4}, UR-DMU \cite{c:10} and Zhang \etal\cite{zhang2022exploiting}. The AP results on XD-Violence dataset are presented in Table \ref{table1}. 

When evaluated on video-level labels for supervision, our approach achieves state-of-the-art performance, surpassing all unsupervised methods by a significant margin in AP. Compared with previous weakly-supervised unimodal methods, our approach achieves a minimum of 4.01\% improvement over their results. When compared with the state-of-the-art weakly-supervised multimodal method, MACIL-SD \cite{c:4}, our approach achieves a substantial improvement of 2.27\%.
These results demonstrate the effectiveness of our proposed method for learning instance representations in hyperbolic space, and its potential for enhancing the performance of violence detection models.

In comparison to other methods, our approach has the smallest model size (0.607M), while still outperforming all previous methods. These results demonstrate the efficiency of our framework, which leverages a simpler network architecture while achieving superior performance. For model complexity and inference power, Table \ref{table5} presents the average inference time and FLOPs (floating point operations) computed on the test set. Due to existing computational toolkits (such as fvcore) did not support some special mathematical functions for FLOPs, so we provide inference time here and find that the inclusion of hyperbolic geometry introduces a slight increase in inference cost for the model but the performance improvement is significant.

\begin{figure*}[t]
	\centering
	\includegraphics[width=0.8\textwidth]{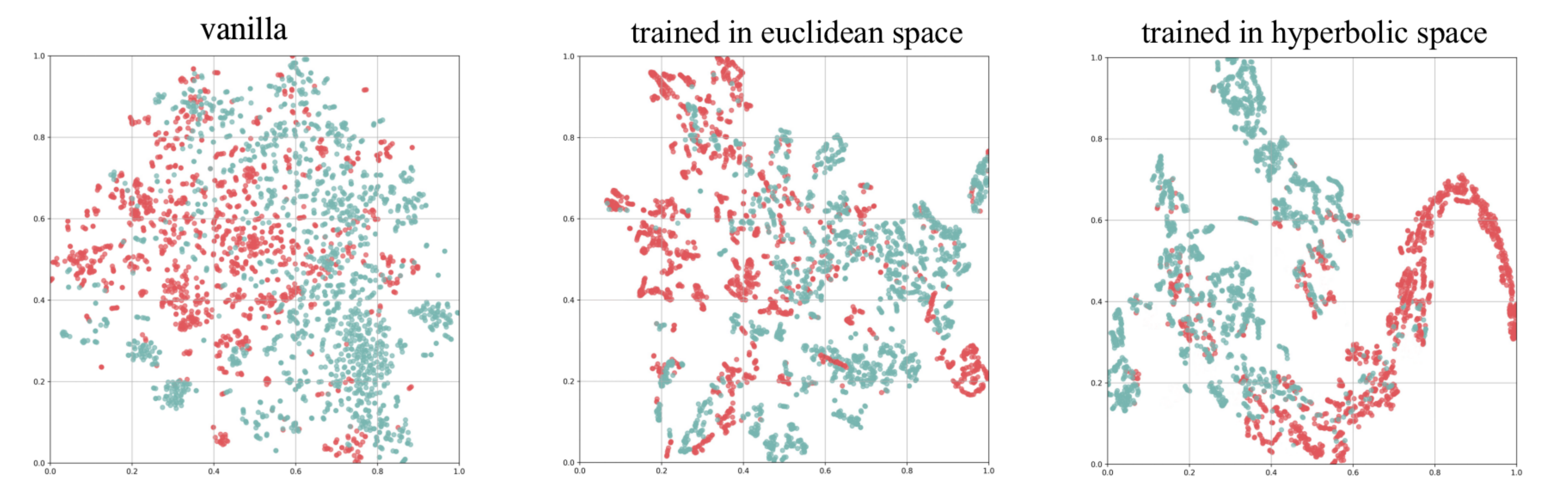} 
	\caption{Feature space visualizations of the vanilla features (left), the trained features via Euclidean space (middle), and trained features via hyperbolic space (right). All the results are performed on XD-Violence test set. Red dots represent non-violent features, and green dots denote violent features.}
	\label{fig4}
\end{figure*}

\begin{figure*}[h]
	\centering
	\includegraphics[width=0.85\textwidth]{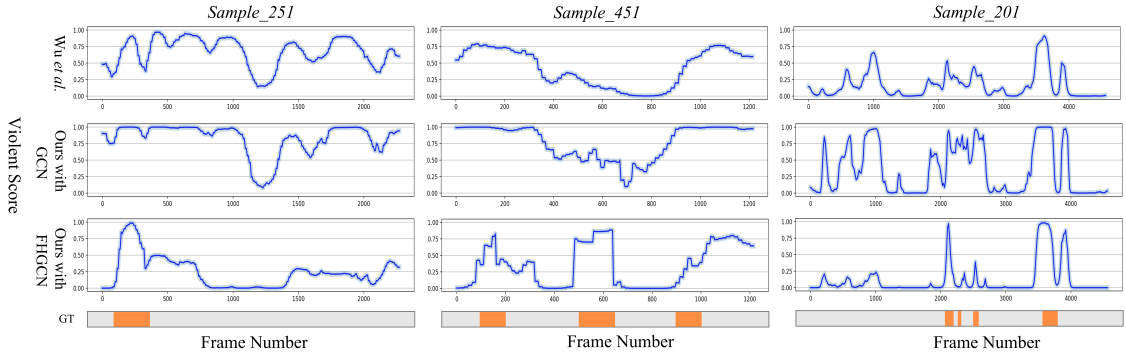} 
	\caption{Ablative visualization of testing results on XD-Violence. The blue curves are predicted violent scores, and the "GT" bars in orange are ground truths of violent regions.}
	\label{fig5}
\end{figure*}

\begin{table}[h]
	\caption{Ablation studies for the proposed Hyperbolic Feature Similarity Graph (HFSG) branch and Hyperbolic Temporal Relation Graph (HTRG) branch.}
    \centering
	\label{table4}
        \resizebox{0.75\columnwidth}{!}{\begin{tabular}{cccc}
		\toprule
		Index  & HFSG Branch & HTRG Branch &  AP(\%)\\
		\midrule
		1 &  - & \checkmark & 80.58\\
		2 & \checkmark & - & 69.01\\
		3 &  \checkmark & \checkmark & \textbf{85.67} \\ 
		\bottomrule
	\end{tabular}}
\end{table}

\vspace{-2mm}
\begin{table}[h]
	\caption{Ablative results of model complexity and inference power. Inference time (Time) is performed for one iteration on the test set with 5 iterations for warm-up.}
    \centering
	\label{table5}
        \resizebox{1.0\columnwidth}{!}{\begin{tabular}{clccc}
		\toprule
		  Index & Method       &  AP(\%)       & FLOPs(G) & Time(s) \\
		\midrule
		1 & Wu \etal         & 78.64         & 25.898   & 2.922\\ \midrule
		2 & GCN     & 79.85         & 17.164   & 2.843\\
		3 & HGCN   & 82.88         & -         & 3.395\\ 
  		4 & FHGCN \& Cosine Similarity   & 83.25         & -           & 2.896 \\ 
            5 & FHGCN \& Lorentz Similarity  & 85.67   & -    & 3.090 \\  
		\bottomrule
	\end{tabular}}
\end{table}

\subsection{Qualitative Results}
To further evaluate our method, we first visualize prediction results on XD-Violence and shown in Figure \ref{fig3}. As exhibited in the figures for violent videos, our method not only produces a precise detection area but also generates higher anomaly scores than normal ones. In non-violent videos, our method produces almost zero predictions for normal snippets. 

In addition, we provide Figure \ref{fig4} to show feature space visualizations of the vanilla, euclidean, and hyperbolic trained features. The hyperbolic features are first transformed into Euclidean space for computation using t-SNE \cite{Maaten_Hinton_2008} . The results demonstrate clear clustering of violent and non-violent features in the hyperbolic space, with increased distances between uncorrelated features after training. Notably, features trained in hyperbolic space needs to be transformed into euclidean space and then computed by the t-SNE tool. We also provide the CO-SNE \cite{guo2022co} visualization designed for hyperbolic space in the Appendix.



\vspace{-2mm}
\subsection{Ablation Studies}
To investigate the contribution of key components in the proposed framework, we further conduct extensive ablation studies to demonstrate its efficiency. 

We first conduct comparative experiments on different multimodal fusion manners, and the results are shown in Table \ref{table2}. Our detour fusion achieves a performance of 85.67\% with a 2.32\% improvement than simply utilizing concatenation (Concat) fusion. Besides, Wu \etal \cite{c:2} adopt a concatenation manner of early fusion strategy. We re-implement their method using our detour fusion module and get the improvement of 1.22\%.

Then we investigate the contribution of Fully Hyperbolic GCN (FHGCN) to our framework with results in Table \ref{table3}, revealing a remarkable performance boost from 76.87\% to 85.67\% compared to standard GCN in Euclidean space. Moreover, the numerical stability of FHGCN equipped with the Lorentz model enabled our method to outperform HGCN with the Poincare model, achieving a 2.79\% improvement. As shown in Table \ref{table3}, we also evaluate the model performance using diverse feature similarity metrics. Our findings demonstrate that using Lorentzian distance for the Lorentz model yields a superior capacity for capturing feature similarity in the hyperbolic space and consequently, it outperforms alternative methods. Besides, the contributions of the proposed HFSG branch and HTRG branch are analyzed. The results in Table \ref{table4} indicate the importance of each branch. When equipped with both branches, our method can achieve the best performance of 85.67\% AP.

Finally, in Figure \ref{fig5}, we showcase prediction results to facilitate qualitative analysis. The visual comparison reveals that our method, leveraging hyperbolic geometry, effectively mitigates predictive noise in both violent and non-violent snippets, surpassing the baseline and variant methods that utilize Euclidean geometry. This demonstrates the exceptional capability of our approach in capturing subtle semantic discrepancies that were previously indistinguishable.


\begin{figure}[h]
	\centering
	{\includegraphics[width=0.47\columnwidth, clip=true, trim=0 0 0 0]{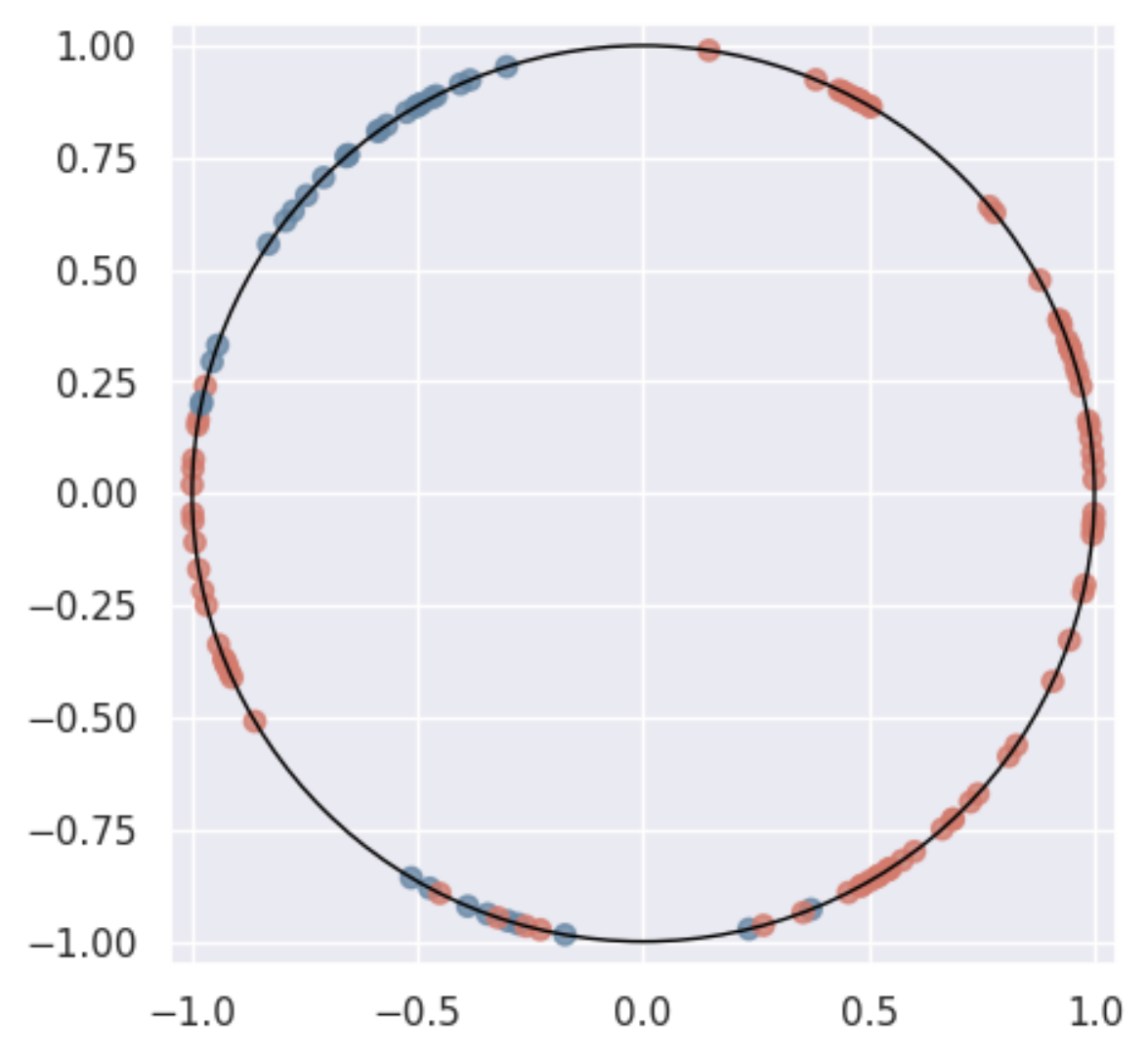}}\vspace{-0.05cm}
	{\includegraphics[width=0.47\columnwidth, clip=true, trim=0 0 0 0]{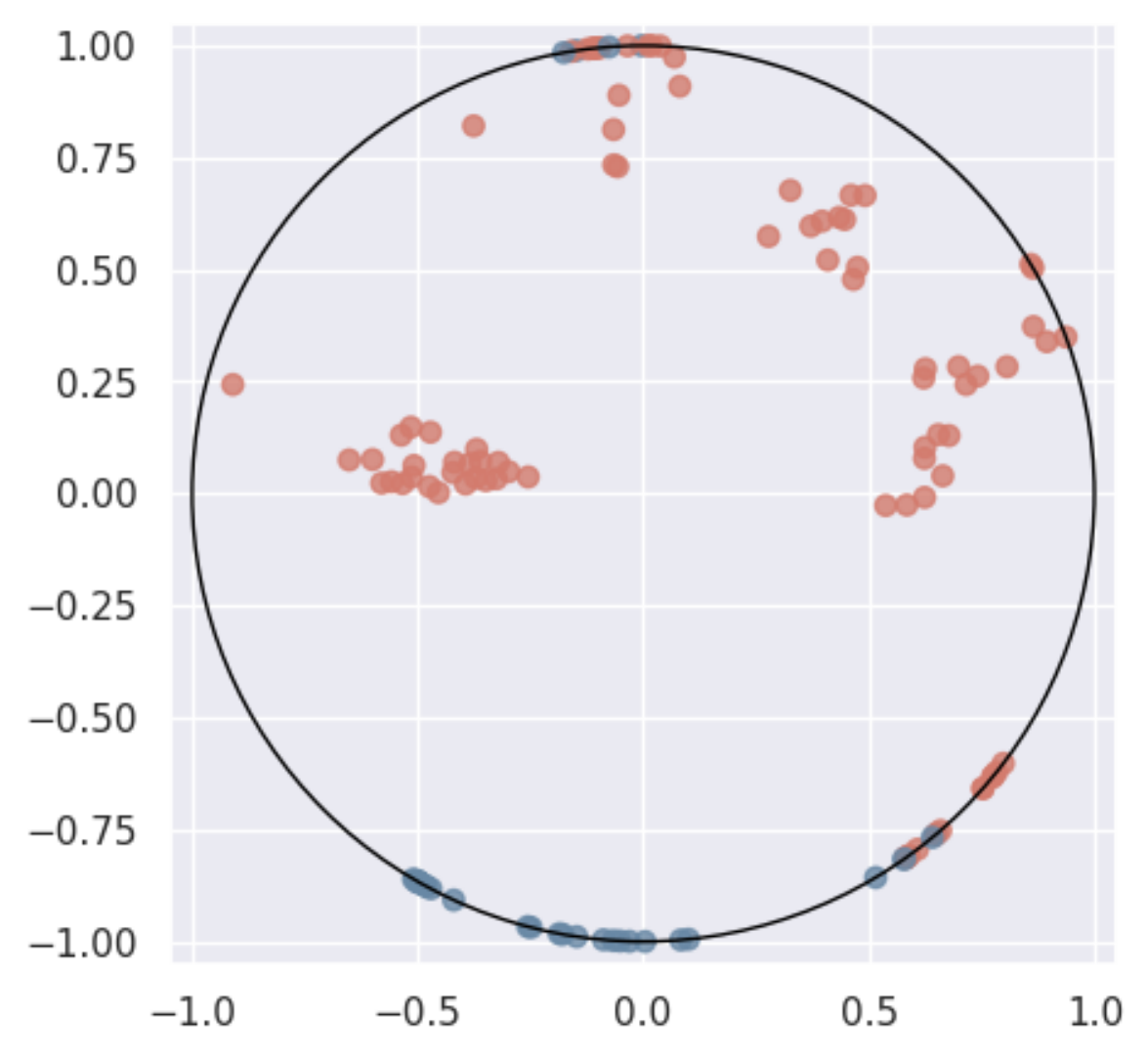}}\vspace{-0.05cm}
	\subfigure[Vanilla Embeddings]{\includegraphics[width=0.47\columnwidth, clip=true, trim=0 0 0 0]{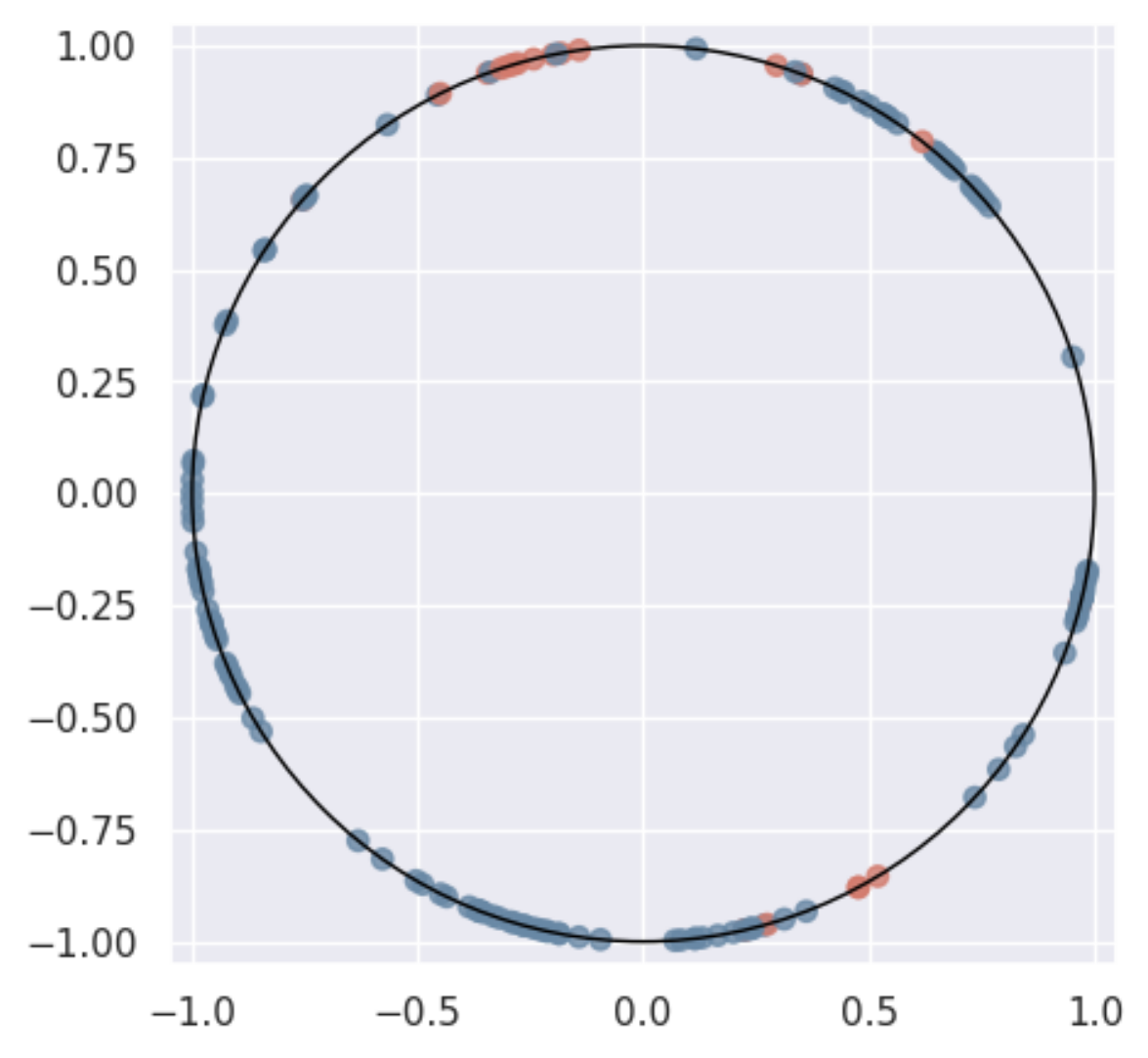}}
	\subfigure[Trained Embeddings]{\includegraphics[width=0.47\columnwidth, clip=true, trim=0 0 0 0]{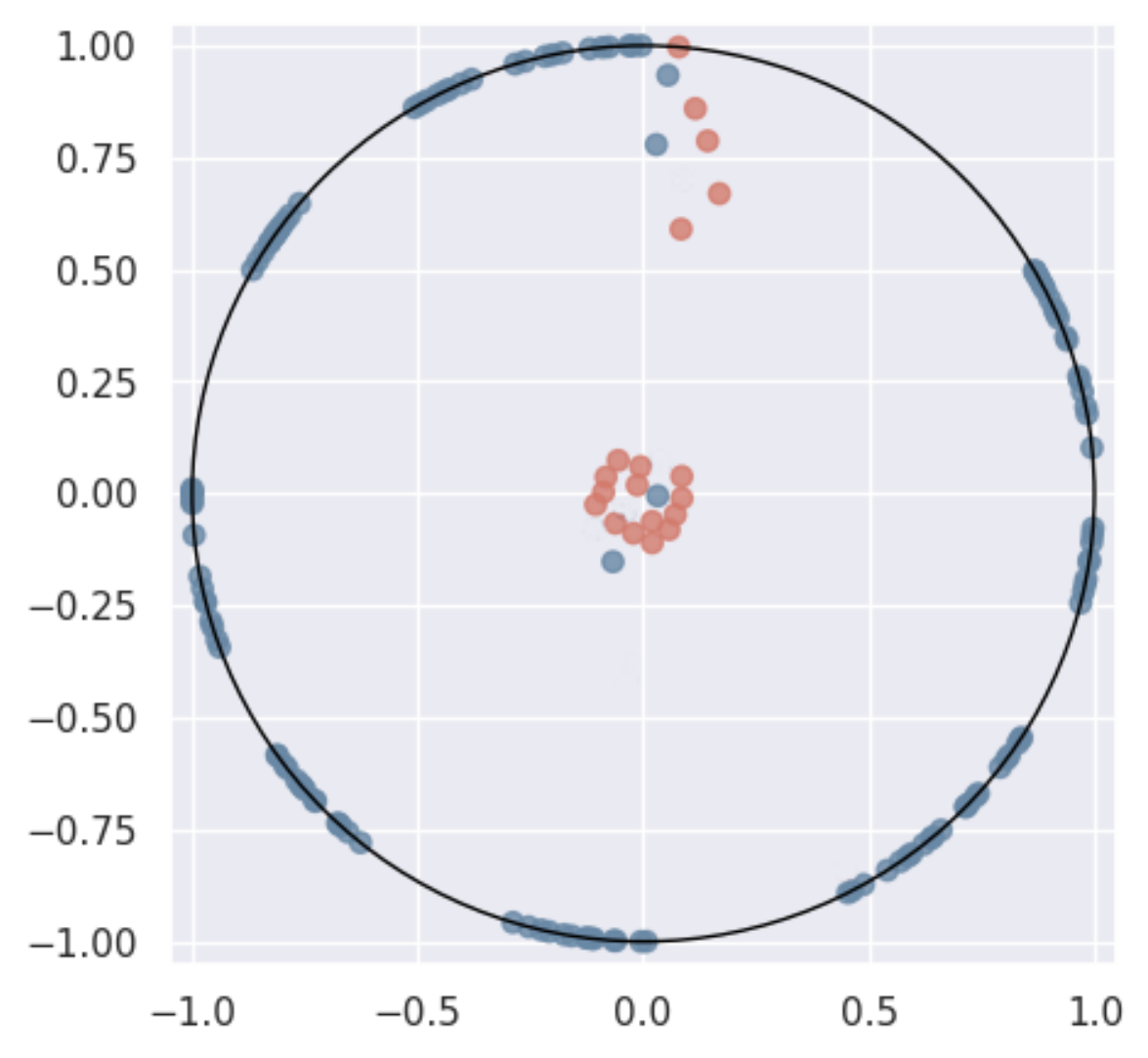}}
	\caption{The projection of high-dimensional vanilla embeddings and output hyperbolic embeddings of our model in a two-dimensional features space with CO-SNE \cite{guo2022co}, which can preserve the hierarchical and similarity structure of the high-dimensional hyperbolic data points. The red points indicate violent embeddings and the blue points indicate non-violent embeddings.}\vspace{-0.4cm}
	\label{fig6}
\end{figure}

\section{Additional Results and Analysis}
\subsection{Complexity Analysis}
Our method is also designed to be computationally efficient, without introducing an excessive number of parameters. The detour fusion module, which learns the visual features by fully-connected layers, contains the primary model parameters. In contrast, the HFSG and HTRG branches are comparatively lightweight, consisting mainly of hyperbolic graph convolution layers that operate on the learned embeddings. In comparison to other methods, our approach has the smallest model size (0.607M), while still outperforming all previous methods. These results demonstrate the efficiency of our framework, which leverages a simpler network architecture while achieving superior performance.

\begin{figure}[t]
	\centering
	\includegraphics[width=0.45\textwidth, clip=true, trim=0 0 0 25]{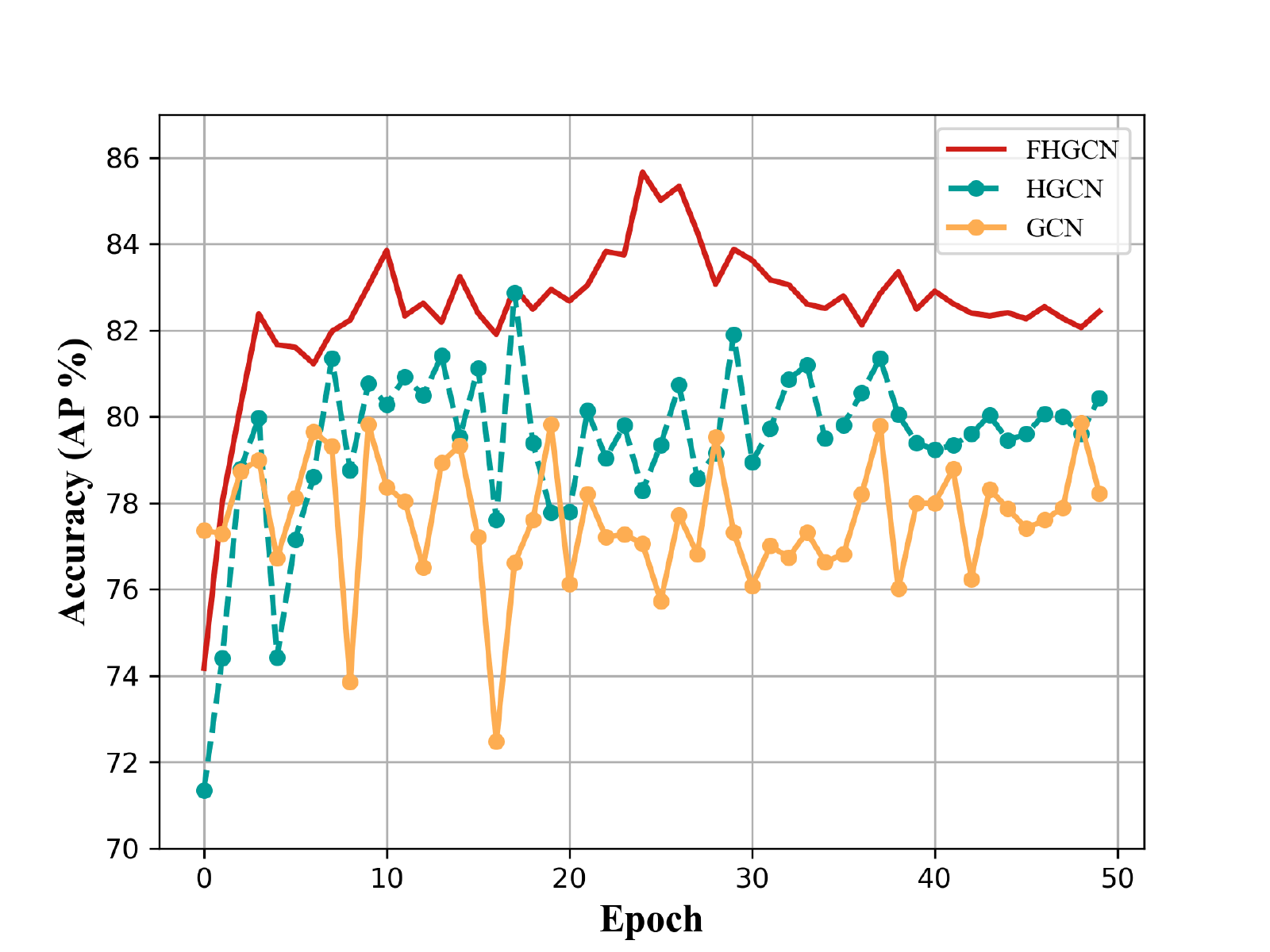} 
	\caption{Comparative results of the accuracy curves in 50 epochs during training.}
	\label{fig7}
\end{figure}

\subsection{Training Stability}
We further provide comparative results of the accuracy curves in 50 epochs during training as shown in Figure \ref{fig7}. Notably, the similarity matrics of hyperbolic feature similarity branch in HGCN and FHGCN are measured by \poincare distance and Lorentzian distance metrics, respectively. As shown, the GCN-based method produces significant jittering results. Thanks to the numerical stability of the Lorentz model, our method that is equipped with FHGCN is more steady compared to other methods during the whole training process.

\subsection{Ablative Results with Different Hyper-parameters}
As illustrated in Table 1 and Table \ref{ap:tab1} and Table \ref{ap:tab2}, we also provide ablative results of different hyper-parameters adopted in our method.  In table 2, compared to Euclidean-based method (such as Wu \etal \cite{Pu_Wu_2022}), the model can obtain promising results (80.46\%) with small embedding dimension (32) and maintain lightweight (0.609M) and fast (2.585s). Table \ref{ap:tab2} illustrates the effects of different hidden dimensions and layers of FHGCN on model performance.

\subsection{CO-SNE and T-SNE Visualization}
We apply CO-SNE \cite{guo2022co} designed for hyperbolic data to visualize the vanilla embeddings and trained embeddings produced by the hyperbolic neural network. For high-dimensional hyperbolic datapoints which are close to the boundary of the \poincare ball, the standard t-SNE often wrongly underestimates the distance between them and would lead to low-dimensional embeddings collapse into one point, resulting in poor visualization\cite{guo2022co}. Specifically, we adopt the transformation function to project the embeddings of the Lorentz model into \poincare space and then utilize CO-SNE for visualization. As shown in Figure \ref{fig6}, where the left column shows vanilla embeddings without training and the right column shows trained embeddings by our model, we can observe that violent and non-violent features are well separated after training, \eg violent features are close to the center while non-violent features are pushed away to the boundary.

\begin{table}[t]
\centering
	\caption{Ablative results of different input dimensions of hyperbolic GCN in our method. Notably, to input any size of input dimension of HFSG and HTRG branches, we adopt a concatenation manner for multimodal fusion. Inference time (Time) is performed for one iteration on test set with 5 iterations for warmup.}
	\label{ap:tab1}
	\resizebox{0.8\columnwidth}{!}{\begin{tabular}{cccc}
		\toprule
                Input Dimension	 & AP(\%)	& Params(M)	& Time(s) \\\midrule
                256 &	83.35	 &0.758	 &3.277 \\
                128 &	82.28	 &0.664	 &2.905 \\
                64	     &81.32	 &0.627	 &2.788 \\
                32	     &80.46  &0.609	 &2.585 \\
		\bottomrule
	\end{tabular}}
\end{table}
\vspace{-1mm}

\begin{table}[t]
\centering
	\caption{Ablative results of different layers and hidden dimensions of hyperbolic GCN in our method. The left three columns are the results of different layers and the right three ones are for different hidden dimensions.}
	\label{ap:tab2}
	\resizebox{1.0\columnwidth}{!}{\begin{tabular}{ccc|ccc}
		\toprule
                Layers	&AP(\%)	&Params(M)	& Hidden Dimension	&AP (\%)	&Params(M)\\ \midrule
                2 (ours)	&85.67	&0.607	& 16	&82.70	&0.599\\
                4	&83.84	&0.611	& 32 (ours)	&85.67	&0.607\\
                6	&82.30	&0.616	& 64	&84.43	&0.616\\
		\bottomrule
	\end{tabular}}
\end{table}
\vspace{-1mm}



\section{Conclusion}
In this paper, we investigate the modality inconsistency under audio-visual scenarios and the weakness of learning instance representations in Euclidean space. Then a HyperVD framework incorporated with a detour fusion module and two hyperbolic graph learning branches is proposed to address the above issues. To be specific, we design a detour fusion strategy to suppress the negative impacts of audio signals to alleviate information inconsistency across modalities. Furthermore, a hyperbolic feature similarity graph branch and a hyperbolic temporal relation graph branch are proposed to learn similar characteristics and temporal relationships among snippets, respectively. Our HyperVD greatly outperforms previous methods on the XD-Violence dataset, demonstrating the superiority of instance representation learning in hyperbolic space. 

We are convinced that hyperbolic geometry holds great potential for various video understanding and interpretation tasks, such as video anomaly detection and event localization. We are committed to further exploring the power of hyperbolic geometry in these and other related areas in the future.

\printcredits




\bibliographystyle{cas-model2-names}

\bibliography{cas-refs}

\newpage
\bio{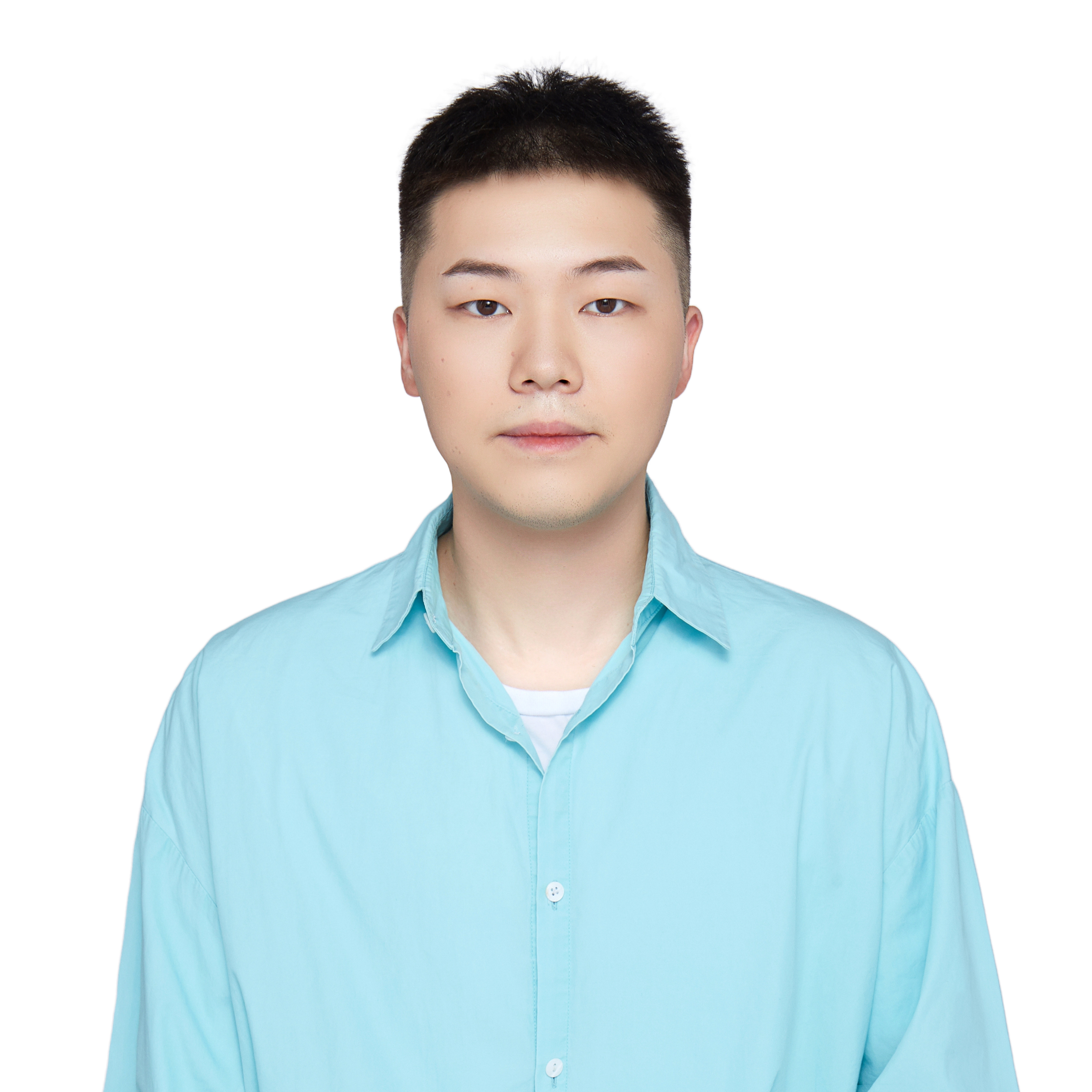}
\textbf{Xiaogang Peng} is currently a postgraduate at the Department of Digital Media Technology, Hangzhou Dianzi University. His research interest includes video understanding, human motion modeling and understanding.
\endbio
\vspace{10pt}

\bio{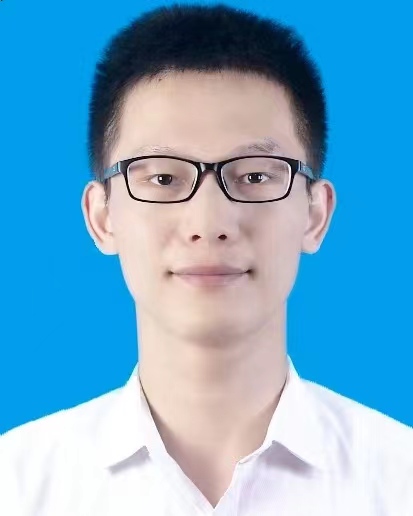}
\textbf{Hao Wen} is currently pursuing the Ph.D. degree in information and communication engineering with the College of Electrical Science and Technology, National University of Defense Technology, Changsha, China. His current research interests include pattern recognition, video understanding and human motion modeling.
\endbio
\vspace{10pt}

\bio{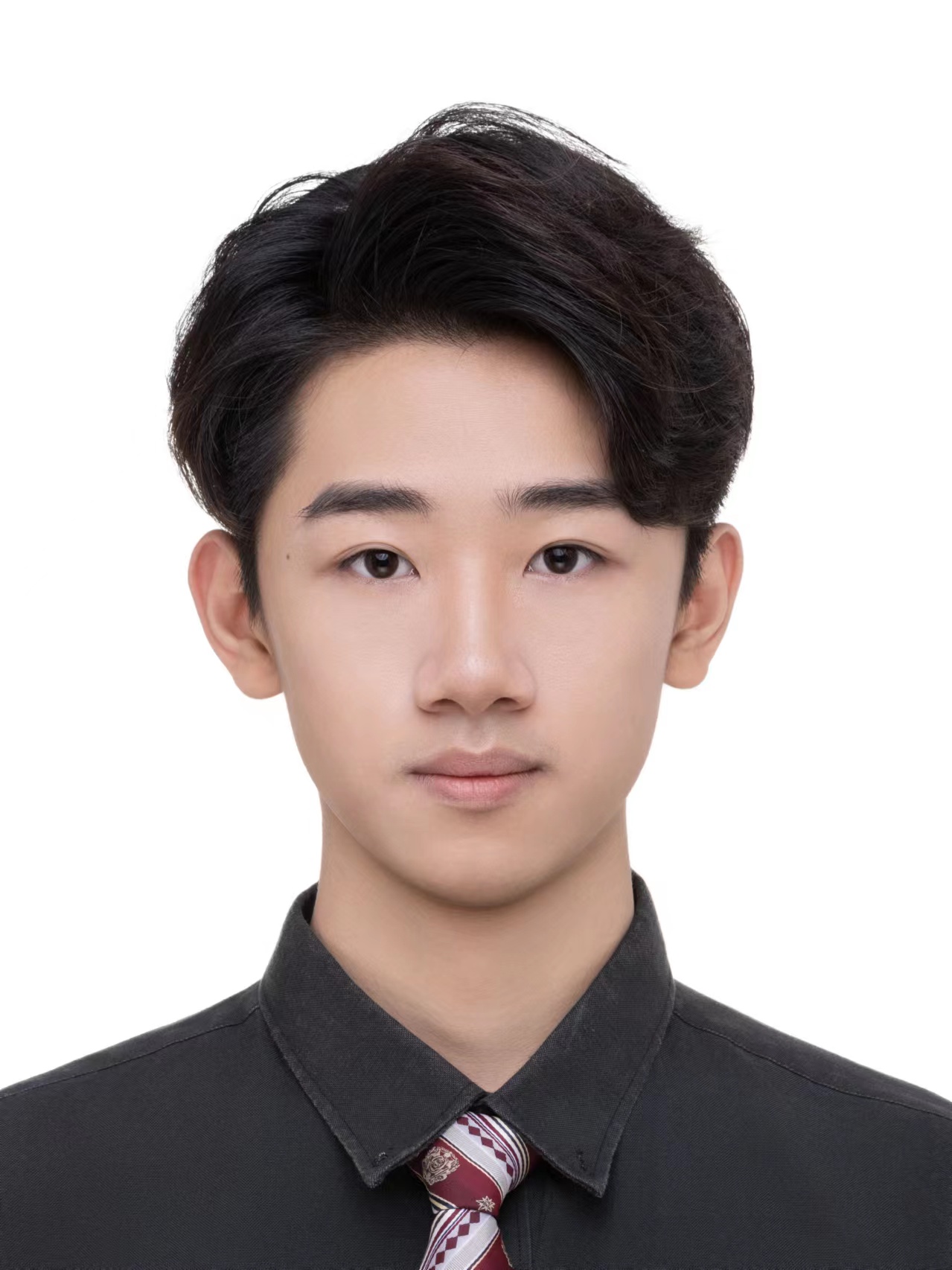}
\textbf{Yikai Luo} is currently an undergraduate student in Digital Media Technology at Hangzhou Dianzi University. His research interest includes Machine vision, and video understanding.
\endbio
\vspace{20pt}

\bio{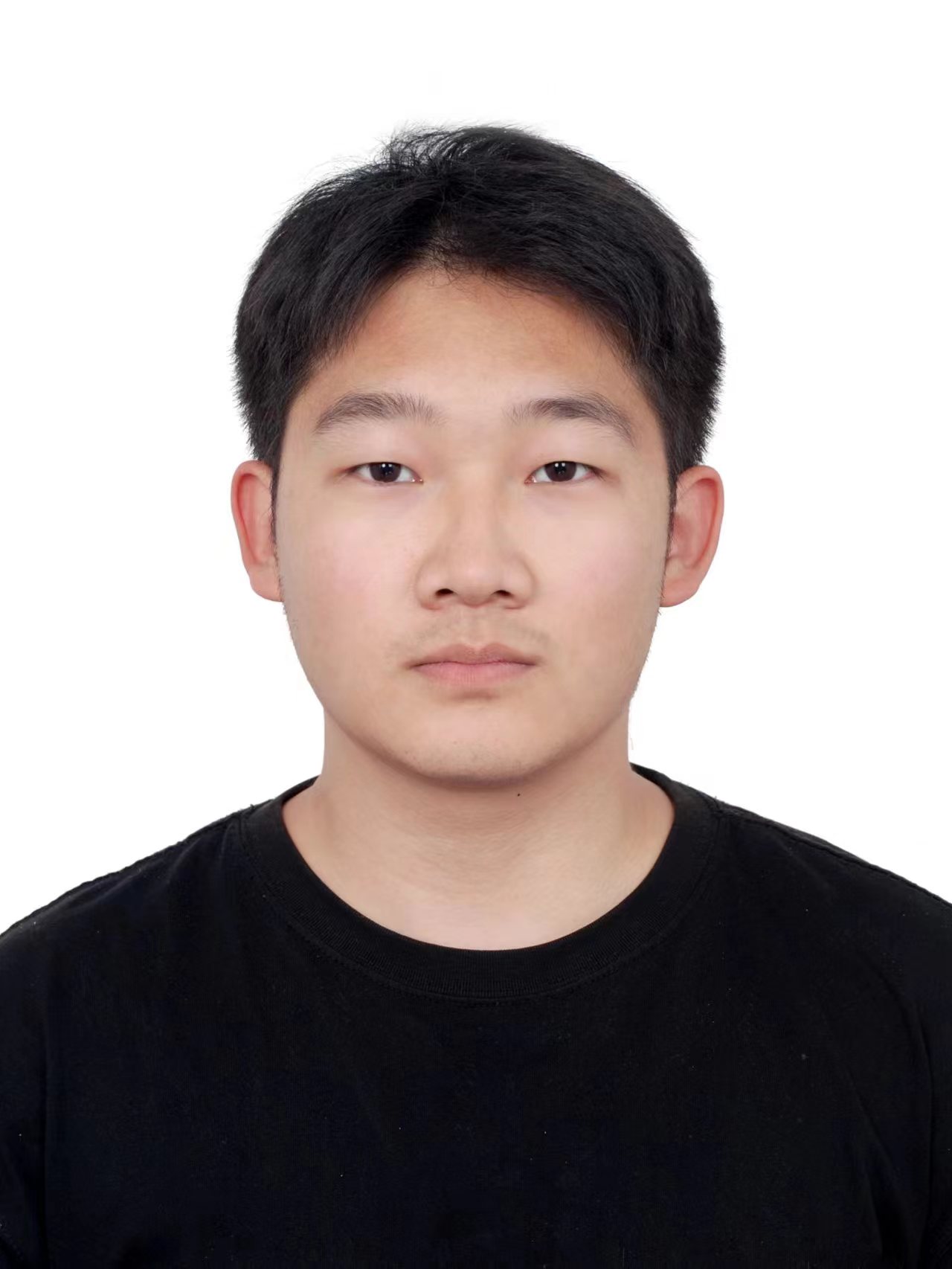}
\textbf{Xiao Zhou} is currently an undergraduate at Hangzhou Dianzi University and majored in Digital Media Technology. His research interest includes human motion modeling and video understanding.
\endbio
\vspace{30pt}

\bio{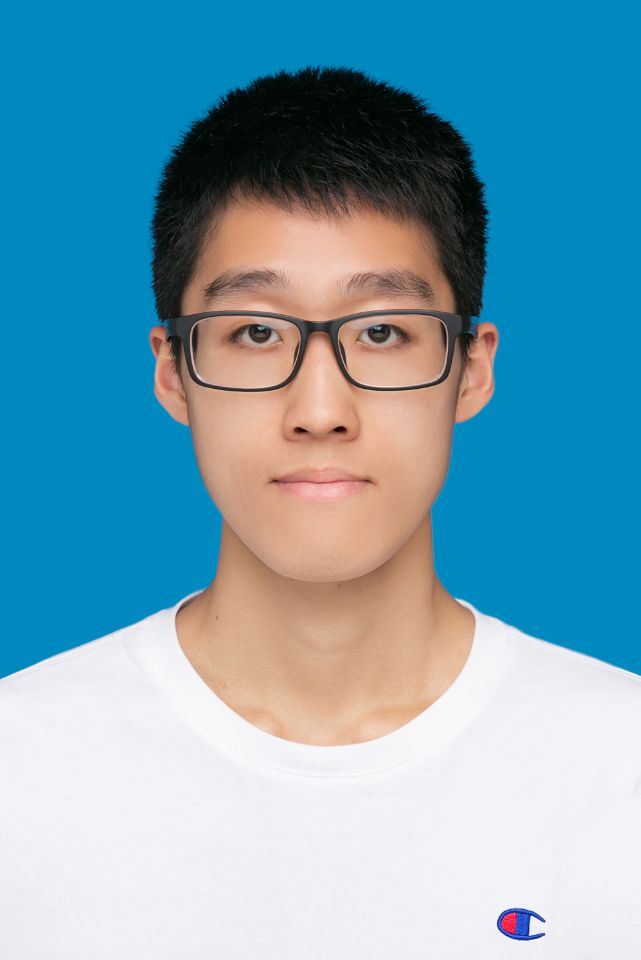}
\textbf{Keyang Yu} is currently an undergraduate at Hangzhou Dianzi University and majored in Digital Media Technology. His research interest includes machine learning and computer vision.
\endbio
\vspace{30pt}

\bio{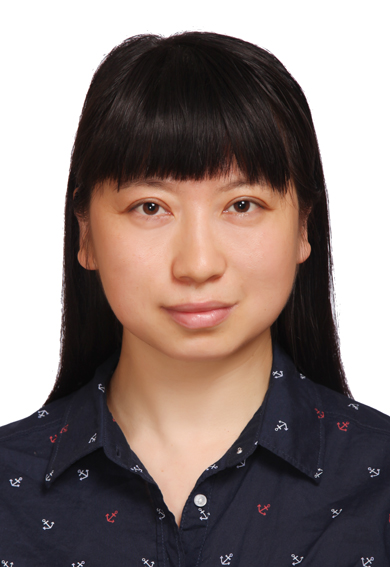}
\textbf{Ping Yang} is currently an Associate Professor with the Faculty of Digital Media Technology, Hangzhou Dianzi University. She received her Ph.D. degree from the School of Optics and Photonics, Beijing Institute of Technology, in 2008. Her research interests include computer vision and color science.
\endbio
\vspace{30pt}

\bio{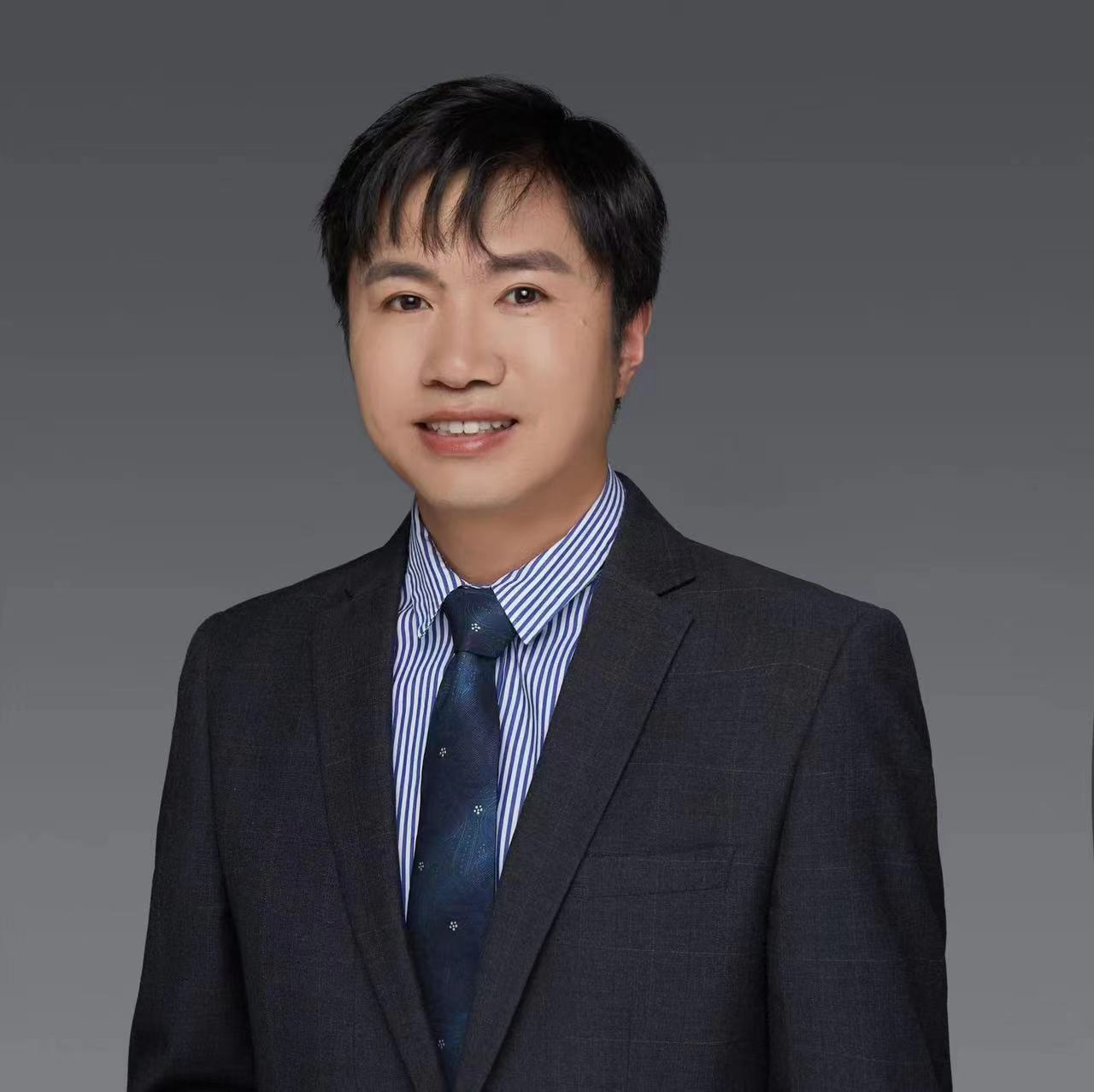}
\textbf{Zizhao Wu} is currently an Associate Professor with the Faculty of Digital Media Technology, Hangzhou Dianzi University. He received his Ph.D. degree from the State Key Laboratory of CAD\&CG, Zhejiang University, in 2013. His research interests include computer vision and computer graphics.
\endbio
\vspace{30pt}

\end{document}